\documentclass[letterpaper,journal]{IEEEtran}
\usepackage{amsmath,amsfonts}
\usepackage{amssymb}
\usepackage{bm}
\usepackage{algorithmic}
\usepackage{algorithm}

\usepackage{array}
\usepackage{pifont}
\usepackage[caption=true,font=normalsize,labelfont=sf,textfont=sf]{subfig}
\usepackage{textcomp}
\usepackage{graphicx}
\usepackage[colorlinks,linkcolor=blue,]{hyperref}
\usepackage{stfloats}
\usepackage{url}
\usepackage{verbatim}
\usepackage{graphicx}
\usepackage{cite}
\usepackage{bm}
\usepackage{xcolor}
\usepackage{multirow}
\usepackage{booktabs}
\usepackage{amsfonts}
\usepackage{enumerate}
\usepackage{siunitx}
\usepackage{gensymb}
\usepackage{bbding}
\usepackage{soul} 
\usepackage{color, xcolor}
\begin{document}

\title{Audio-Visual Class-Incremental Learning for Fish Feeding Intensity Assessment in Aquaculture
}

\author{Meng Cui$^{1}$, Xianghu Yue$^{2}$, Xinyuan Qian$^{3}$, Jinzheng Zhao$^{1}$, Haohe Liu$^{1}$, \\ Xubo Liu$^{1}$, Daoliang Li$^{4}$, Wenwu Wang$^{1}$

\thanks{M. Cui, J. Zhao, H. Liu, X. Liu, and W. Wang are with the Centre for Vision, Speech and Signal Processing (CVSSP), University of Surrey, Guildford GU2 7XH, UK. (e-mail:
[m.cui, j.zhao, haohe.liu, xubo.liu,   w.wang]@surrey.ac.uk).}
\thanks{X. Yue is with the College of Intelligence and Computing, Tianjin University, Tianjin, China. (e-mail:  xianghu.yue@u.nus.edu).}
\thanks{X. Qian is with the School of Computer and Communication Engineering, University of Science and Technology Beijing, Beijing, China (e-mail:  qianxy@ustb.edu.cn).}
\thanks{D. Li is with the National Innovation Center for Digital Fishery, China Agricultural University, China (e-mail: dliangl@cau.edu.cn).}
}

\maketitle
\begin{abstract}

Fish Feeding Intensity Assessment (FFIA) is crucial in industrial aquaculture management. Recent multi-modal approaches have shown promise in improving FFIA robustness and efficiency. However, these methods face significant challenges when adapting to new fish species or environments due to catastrophic forgetting and the lack of suitable datasets. To address these limitations, we first
introduce \textit{AV-CIL-FFIA}, a new dataset comprising 81,932 labelled audio-visual clips capturing feeding intensities across six different fish species in real aquaculture environments. 
Then, we pioneer audio-visual class incremental learning (CIL) for FFIA and demonstrate through benchmarking on AV-CIL-FFIA that it significantly outperforms single-modality methods. 
Existing CIL methods rely heavily on historical data. Exemplar-based approaches store raw samples, creating storage challenges, while exemplar-free methods avoid data storage but struggle to distinguish subtle feeding intensity variations across different fish species.
To overcome these limitations, we introduce \textit{HAIL-FFIA}, a novel audio-visual class-incremental learning framework that bridges this gap with a prototype-based approach that achieves exemplar-free efficiency while preserving essential knowledge through compact feature representations. Specifically, HAIL-FFIA employs hierarchical representation learning with a dual-path knowledge preservation mechanism that separates general intensity knowledge from fish-specific characteristics. Additionally, it features a dynamic modality balancing system that adaptively adjusts the importance of audio versus visual information based on feeding behaviour stages.
Experimental results show that HAIL-FFIA is superior to exemplar-based and exemplar-free methods on AV-CIL-FFIA, achieving higher accuracy with lower storage needs while effectively mitigating catastrophic forgetting in incremental fish species learning.
This work highlights the potential of audio-visual CIL in enhancing the adaptability and robustness of aquaculture monitoring systems. 
Code and data are available at \url{https://github.com/FishMaster93/AV-CIL-FFIA}.

\end{abstract}

\section{Introduction}

\IEEEPARstart
{F}{ish} Feeding Intensity Assessment (FFIA) plays a crucial role in digital aquaculture, significantly enhancing operational efficiency, sustainability, and productivity \cite{cui2024multimodal, atoum2014automatic}. An accurate FFIA enables farmers to precisely monitor fish appetite levels and optimise feeding strategies during the feeding process, reducing waste and environmental impact while improving fish health and growth rates in commercial aquaculture systems \cite{li2020automatic, cui2022fish}.

The traditional audio- and visual-based machine learning approach defines fish feeding intensity as a classification task (e.g., ``\textit{None}", ``\textit{Weak}", ``\textit{Medium}" and ``\textit{Strong}" \cite{du2023feeding, ubina2021evaluating}). Audio data captures subtle feeding sounds that may be invisible to cameras, while visual data provides spatial context that might be ambiguous in audio-only data \cite{cui2022fish, li2024review}. However, acoustics-based methods can be limited in capturing the physical characteristics related to fish behaviour, and are prone to underwater acoustic noises, while visual-based methods are generally susceptible to variations in lighting conditions and noise from water surface reflections. Audio-visual methods can overcome these individual limitations by integrating information from both modalities, improving detection accuracy and robustness under varying environmental conditions \cite{du2024harnessing}. However, existing audio-visual FFIA methods are generally optimised for specific species of fish and environmental conditions \cite{cui2025fish}.
When introduced to new fish species or changing environmental conditions, the performance of these systems can be limited. To address this issue, the system needs to be either re-trained on all historical and new data (often impractical in aquaculture deployments), or fine-tuned only on new species \cite{van2024continual}, which may suffer from the problem of ``\textit{catastrophic forgetting}''\cite{lee2017overcoming} - the loss of previously acquired knowledge when adapting to new data. This limitation restricts the scalability of FFIA technology in diverse settings. Class-incremental learning (CIL) approaches \cite{zhou2023deep}, which enable continuous learning while preserving existing knowledge, offer a potential solution to this challenge.

CIL approaches generally fall into two categories: exemplar-based methods \cite{chen2023dynamic, douillard2020podnet,liu2021adaptive} that store representative samples from previous classes, and exemplar-free methods \cite{meng2024diffclass, petit2023fetril, goswami2023fecam, li2017learning} that preserve knowledge without retaining raw data. In FFIA applications, exemplar-based methods may not be a promising choice, as storing high-dimensional audio-visual data becomes problematic, especially for commercial aquaculture systems with resource constraints, posing limitations in their scalability to the number of fish species and data volume.   

In contrast, exemplar-free methods employ parameter regularization \cite{luo2023representation, shi2023multi}, knowledge distillation \cite{szatkowski2024adapt}, or architectural modifications \cite{sun2023exemplar} to preserve previous knowledge without storing raw data \cite{zhou2024class,wang2022learning}. However, exemplar-free approaches often struggle to maintain consistent performance between different species of fish in real-world aquaculture settings, where environmental variations and species-specific behaviours have a significant impact on feeding intensity patterns \cite{zhu2021prototype}. Prototype-based methods such as prototype calibration \cite{wang2023few, zhang2023few} and enhancement \cite{zhu2021prototype, li2023class,han2023prototype} have emerged as middle-ground solutions that store representative feature vectors instead of raw data. However, these existing prototype approaches have not been adequately explored in multimodal scenarios such as FFIA, where cross-modal interactions and species-specific variations present unique challenges. This gap becomes particularly significant when considering the substantial benefits that multimodal approaches offer for complex environmental monitoring tasks.

Recent advances in audio-visual CIL have demonstrated that multimodal approaches consistently outperform single-modality systems in complex environments \cite{mo2023class, pian2023audio,you2023incremental, yue2024mmal}. 
 Despite these promising results, research on audio-visual incremental learning for aquaculture applications remains unexplored, creating a significant gap for FFIA systems. \textcolor{black}{This gap is particularly challenging because fish species exhibit both shared and species-specific patterns in their feeding behaviours. While general intensity levels (None, Weak, Medium, Strong) share common characteristics in species, each species also displays unique audio-visual signatures during feeding. Traditional CIL approaches struggle with this hierarchical knowledge structure, either overgeneralizing species or failing to leverage common patterns.}
To address these challenges, we propose a \textbf{H}ierarchical \textbf{A}udio-visual \textbf{I}ncremental \textbf{L}earning for \textbf{FFIA} (\textbf{\textit{HAIL-FFIA}}). HAIL-FFIA is a novel hierarchical audio-visual CIL framework specifically designed for FFIA.
Building upon recent advances in prototype-based methods, our approach employs an exemplar-free strategy that stores only clustered feature prototypes rather than raw data. Unlike existing prototype approaches, HAIL-FFIA features a unique dual-path architecture that separates general intensity knowledge from fish-specific characteristics, enabling effective cross-species knowledge transfer while dynamically balancing audio and visual information based on feeding behaviour stages. Moreover, to facilitate research in this emerging area, we introduce \textbf{\textit{AV-CIL-FFIA}}, the first comprehensive dataset designed for audio-visual class-incremental learning in FFIA, comprising \textbf{81,932} labelled audio and video clips across six different fish species in real aquaculture environments. Our key contributions can be summarised as follows.

\begin{itemize}
    \item We propose HAIL-FFIA, the first hierarchical audio-visual class-incremental learning framework specifically designed for fish feeding intensity assessment, which implements an exemplar-free prototype-based approach to address the problem of catastrophic forgetting.
    \item We introduce a dual-path architecture that uniquely separates general intensity knowledge from fish-specific characteristics while developing a dynamic modality balancing mechanism that adaptively adjusts the importance of audio versus visual information based on feeding behaviour stages, enabling effective cross-species knowledge transfer with minimal storage requirements.
    \item We present AV-CIL-FFIA, the first dataset designed for audio-visual class-incremental learning in FFIA, comprising 81,932 labelled clips across six fish species in real aquaculture environments.
    \item We establish the first comprehensive benchmarks for incremental learning in FFIA, evaluating both single and multi-modal approaches to provide valuable baselines for future research in this emerging field.

\end{itemize}

The remainder of this paper is organised as follows: Section \ref{sec: related work} reviews related work.  Section \ref{sec: AV-CIL} describes our proposed method, and Section \ref{sec:dataset} introduces the new audio-visual dataset designed for the CIL-based FFIA.  Experimental details are shown in Section \ref{sec:experiment}. Section \ref{sec:results} presents our experimental results and discussion, and Section \ref{sec:conclusion} concludes with a discussion on future research directions.

\section{Related work}
\label{sec: related work}

\subsection{Audio-visual FFIA}
Audio-visual learning leverages the complementary nature of these two modalities to achieve a more robust and comprehensive understanding of scenes and events \cite{zhu2021deep, wei2022learning}. The integration of audio and visual modalities for FFIA has emerged as a promising approach to overcome the limitations of single-modality systems \cite{gu2025mmfinet, du2024harnessing}. Multimodal fusion captured both the acoustic signatures of feeding activities and their corresponding visual manifestations, achieving a 15-20\% improvement in accuracy compared to the best single-modality based alternatives \cite{cui2024multimodal, yang2024fish}. The audio modality excels at detecting subtle feeding sounds even in turbid water, while visual data provide spatial context and behavioural patterns that may be acoustically ambiguous. Du et al. \cite{du2024harnessing} further established that audio-visual FFIA systems maintain robust performance under various environmental challenges. Their work showed that when visual data quality was degraded due to poor lighting or water turbidity, the audio modality was effective in making compensations, and vice versa, when acoustic interference was present \cite{zhang2023intelligent}. This cross-modal compensation mechanism enables a more reliable feeding assessment in the diverse conditions encountered in commercial aquaculture settings. 

Despite these advances, existing audio-visual FFIA approaches have primarily been developed and evaluated within static settings involving fixed fish species. Most systems assume a closed-world scenario in which the fish species remain constant over time, requiring complete retraining when new species are introduced. This limitation becomes particularly problematic in commercial aquaculture, where operations frequently expand to include new species or transfer existing monitoring systems to different environments. The challenge of adapting audio-visual FFIA systems to new fish species while maintaining performance on previously learned ones represents a significant gap in current research, highlighting the need for incremental learning approaches specifically designed for multimodal fish feeding assessment.


\subsection{Class-Incremental Learning}
Class-Incremental Learning (CIL) aims to enable models to continuously learn from new classes while preserving knowledge of previously learned classes. This capability is crucial for FFIA systems that must adapt to new fish species without forgetting previously learned ones. Current CIL approaches can be broadly categorised into exemplar-based and exemplar-free methods.

\subsubsection{Exemplar-based methods}

Exemplar-based methods store and utilise data from previous learning stages to mitigate catastrophic forgetting. These approaches include memory replay \cite{wang2022memory, li2024adaer}, which revisits stored examples during training on new classes; distillation-based methods, which transfer knowledge from old to new models using stored samples \cite{rebuffi2017icarl, he2018exemplar,douillard2020podnet}; and dynamic architecture approaches that hold incremental modules to increase the capacity of the model to handle new classes \cite{yan2021dynamically, yoon2017lifelong}. While these methods achieve state-of-the-art performance, they face challenges in FFIA applications due to the storage requirements for high-dimensional audio-visual data and potential deployment constraints in aquaculture monitoring systems.

\subsubsection{Exemplar-free methods}

Exemplar-free approaches address catastrophic forgetting without storing raw data from previous classes, making them suitable for resource-constrained deployments and scenarios where data privacy is a concern \cite{petit2024analysis}. By avoiding the storage of actual examples, these methods eliminate the risk of sensitive data leakage while typically requiring less memory. These methods can be further divided into several categories:

\textbf{Regularization-based methods} constrain parameter updates to preserve previous knowledge. These approaches typically introduce additional terms to the loss function to prevent drastic changes to important parameters \cite{li2024continual}. Elastic Weight Consolidation (EWC) \cite{Kirkpatrick2017overcoming} employs Fisher information matrices to identify and protect the crucial weights for previously learned tasks. Learning without forgetting (LwF) \cite{li2017learning} and similar knowledge distillation techniques act as implicit regularizers by forcing new models to produce outputs similar to previous models on new data, effectively preserving knowledge without storing examples. Other methods like Memory Aware Synapses (MAS) \cite{aljundi2018memory} estimate parameter importance through gradient magnitudes, while Rotated Elastic Weight
Consolidation (R-EWC) \cite{liu2018rotate} adapts the regularization strength based on task similarity.

\textbf{Analytical learning-based methods} replace iterative optimisation with closed-form solutions, offering significant computational efficiency. Analytic Class-Incremental Learning (ACIL) \cite{zhuang2022acil} reformulates incremental learning as a recursive least squares problem, enabling direct parameter updates without accessing historical data by maintaining a correlation matrix. Dual-Stream Analytic Learning (DS-AL) \cite{zhuang2024ds} employs a dual-stream architecture to enhance model fitting capacity while preserving the computational benefits of analytical solutions. Gaussian Kernel Embedded Analytic Learning (GKEAL) \cite{zhuang2023gkeal} introduces Gaussian kernel processes to improve the representation capacity in few-shot scenarios. These approaches are particularly promising for continuous learning applications as they avoid the need for multiple training epochs and provide mathematically guaranteed solutions that minimise catastrophic forgetting.

\textbf{Prototype-based methods} maintain compact class representations in feature space instead of storing raw examples \cite{liu2025class}. These approaches typically preserve class prototypes or centroids that capture essential class characteristics. Prototype Augmentation and Self-Supervision (PASS) \cite{zhu2021prototype} maintains class prototypes and employs prototype augmentation to improve discrimination between old and new classes. Feature Translation for Exemplar-Free Class-Incremental Learning (FeTrIL) \cite{petit2023fetril} generates pseudo-features based on class statistics to enhance training without storing the original data. Continual Prototype Evolution (CoPE) \cite{de2021continual} stores and leverages class prototypes with contrastive learning to maintain discriminative boundaries between classes. These methods offer a balance between memory efficiency and performance by storing only the essential information needed to represent class boundaries, making them particularly suitable for long-term incremental learning scenarios.

\subsection{Research Gap and Our Approach}
While both audio-visual learning and CIL have advanced significantly in recent years, their integration remains largely unexplored, particularly for aquaculture applications. Current CIL methods have focused mainly on tasks based on single modalities \cite{zhou2024class, masana2022class}, without addressing the unique challenges of multimodal data streams in underwater environments. \textcolor{black}{A critical challenge in FFIA is that feeding behaviours exhibit both universal intensity patterns shared across species and species-specific characteristics that vary significantly. Existing approaches fail to model this structured knowledge effectively—either missing opportunities for cross-species knowledge transfer or losing discriminative features. Commercial aquaculture operations need systems that can adapt to new species while maintaining performance on previously learned ones. Our work addresses this gap by proposing HAIL-FFIA, a hierarchical audio-visual CIL framework that explicitly separates general-intensity knowledge from species-specific representations, enabling efficient knowledge transfer while preserving distinctive behaviours.}

\section{Methodology}
\label{sec: AV-CIL}

\subsection{Preliminaries}

CIL for FFIA aims to train a model $M_\theta$ with parameters $\theta$ that can sequentially learn to recognise feeding intensities across different fish species without forgetting previously learned species. In our setting, we define a sequence of $K$ incremental tasks $\{T_1, T_2, \ldots, T_K\}$, where each task represents learning feeding intensity assessment for a new fish species. For an incremental task $T_k$ representing the $k$-th fish species, its corresponding training set can be denoted as $D_k^{\text{train}} \sim\{X_{k,a}^{\text{train}}, X_{k,v}^{\text{train}}, Y_k^{\text{train}}\}$, where $X_{k,a}$ and $X_{k,v}$ are the audio and visual recordings of feeding events in $D_k$, respectively, and $Y_k \in I$ is the corresponding feeding intensity label from a fixed set of intensity categories $C = \{$``\textit{None}", ``\textit{Weak}", ``\textit{Medium}", ``\textit{Strong}"$\}$. Unlike traditional CIL, where the class set expands with each task, in our FFIA setting, the label space $C$ remains constant across all tasks, but each task involves a different fish species with unique audio-visual characteristics when exhibiting the same feeding intensities. The objective of audio-visual CIL for FFIA at incremental step $k$ is to train the model using data from the current fish species $D_k^{\text{train}}$, and evaluate it on $D_{1:k}^{\text{test}}$, which contains test data from all previously seen fish species. Importantly, our framework operates under exemplar-free constraints, meaning that raw audio-visual data from previous fish species cannot be stored or accessed during the training of task $T_k$, making the learning process more challenging but also more practical for deployment in resource-constrained aquaculture settings.

\begin{figure*}
\centering
\includegraphics[scale=0.5]{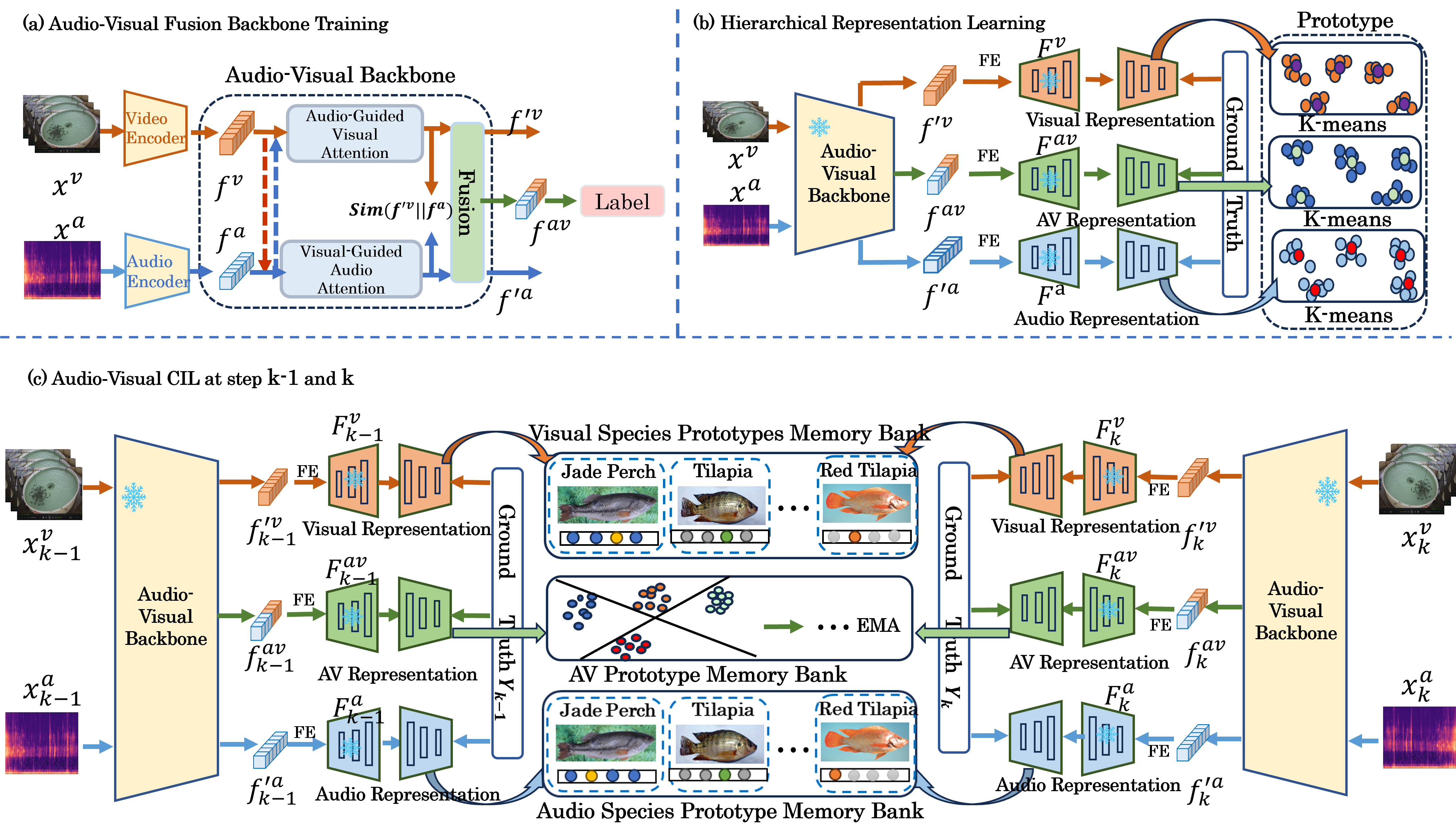}
\caption{The proposed Audio-Visual Class-Incremental learning framework. (a) Audio-Visual Fusion Backbone Training uses cross-modal attention to integrate complementary features from both modalities. (b) Hierarchical Representation Learning separates general feeding intensity knowledge from species-specific features, creating distinct prototype banks. FE(Feature Expansion) increases the representational capacity of the model through a linear transformation that projects features into a higher-dimensional space. (c) During incremental learning at steps $k-1$ and $k$, the framework preserves knowledge through prototype memory banks while adapting to new species without storing raw examples. EMA (Exponential Moving Average) refers to our prototype updating mechanism that uses a weighted average of previous and new prototypes, allowing gradual adaptation to new fish species while preserving knowledge of previous ones.}
\label{fig:AV_Class}
\end{figure*}

\subsection{Hierarchical Audio-visual Incremental Learning for FFIA}
In this subsection, we present our HAIL-FFIA framework, as shown in Fig. \ref{fig:AV_Class}, which employs a novel hierarchical representation approach to address the unique challenges of fish feeding intensity assessment in a class-incremental learning context. Our framework consists of three key components: (1) a dual-encoder feature extraction module, (2) a hierarchical representation learning structure, and (3) a dynamic modality balancing mechanism.

\subsubsection{Dual-Encoder Feature Extraction}
Given an input pair of synchronised visual frames $X^v$ and audio signal $X^a$ capturing a fish feeding event, we first employ pre-trained encoders to extract modality-specific features. We use a pre-trained S3D \cite{xie2018rethinking} as a visual encoder $E^v$ to capture the temporal dynamics of fish movement and feeding behaviour from visual data. For audio data, we employ pre-trained PANNs (MobileNetV2 \cite{kong2020panns}) as an audio encoder $E^a$ to extract spectral features of different feeding intensities. This results in high-dimensional feature vectors $f^v = E^v(X^v)\in \mathbb{R}^{L \times S \times d}$ and $f^a = E^a(X^a)\in \mathbb{R}^{d}$, where $L$ represents temporal frames, $S$ represents spatial locations, and $d$ is the feature dimension.

Given the balanced discriminative power of both visual and acoustic modalities in FFIA, we implement a bidirectional cross-modal attention mechanism that enables each modality to enhance the other. Unlike previous approaches such as those by \cite{pian2023audio, yue2024mmal}, which designate one modality to guide the other, our method treats both modalities as equally valuable information sources. This bidirectional approach adaptively compensates for environmental variations that affect different fish species. For example, visual performance deteriorates in turbid water and insufficient light, while acoustic detection suffers in noisy environments or during quiet feeding behaviours. Our mechanism dynamically adjusts to these changing conditions, maintaining robust performance in such diverse scenarios.

First, we compute bidirectional attention scores:
\begin{equation}
    \begin{split}
        Score^a &= \sigma(f^a W^a), \\
        Score^v_l &= \sigma(f^v_l W^v),
    \end{split}
\end{equation}
where $W^a, W^v \in \mathbb{R}^{d \times d}$ are learnable projection matrices,  $f^v_l$ is the feature from each temporal frame, $Score^a$ is the attention score derived from audio features, $Score^v_l$ is the corresponding attention score for visual frame $l$ and $\sigma(\cdot)$ is a Tanh activation function.

For the spatial-temporal attention in the visual stream, we compute:
\begin{equation}
    \begin{split}
        w^{Spa.}_l &= \text{Softmax}(Score^a \odot Score^v_l), \\
        Score'^v_l &= \sum(w^{Spa.}_l \odot Score^v_l), \\
        w^{Tem.} &= \text{Softmax}([Score'^v_1,...,Score'^v_L]),
    \end{split}
\end{equation}
where $w^{Spa.}_l$ represents spatial attention weights highlighting important regions in frame $l$,  $Score'^v_l$ is the spatially-weighted representation of frame $l$ after applying these weights, $w^{Tem.}$ denotes temporal attention weights that identify the most informative frames across the entire sequence, and $\odot$ denotes the Hadamard (i.e, element-wise) product. Similarly, we compute audio attention guided by visual features:
\begin{equation}
    \begin{split}
        w^{Audio} &= \text{Softmax}(\frac{1}{L}\sum_{l=1}^{L}Score^v_l \odot Score^a),
    \end{split}
\end{equation}
where the averaged visual attention across all frames guides which elements of the audio representation should be emphasised, allowing visual information to enhance audio features.

The enhanced features from both modalities are then computed as:
\begin{equation}
    \begin{split}
        f'^v &= \sum^L_{l=1}w^{Tem.}_l \odot \sum(f^v_l \odot w^{Spa.}_l), \\
        f'^a &= f^a \odot w^{Audio},
    \end{split}
\end{equation}
where $f'^v$ represents visual features enhanced through hierarchical spatial-temporal attention, and $f'^a$ represents audio features enhanced through visually guided attention.

To ensure cross-modal consistency, we introduce a bidirectional similarity constraint:
\begin{equation}
    \mathcal{L}_{sim} = 1 - \frac{f'^v \cdot f'^a}{\|f'^v\| \cdot \|f'^a\|}
\end{equation}
This constraint addresses the inherent domain gap between visual and audio representations. Despite mutual guidance through attention, these modalities exhibit different distributional properties resulting from their distinct physical characteristics. By minimizing the cosine distance, we align these representations in a common semantic space while preserving their complementary information, enhancing robustness when environmental factors degrade either modality.

The initial fused representation integrates both enhanced features with a balanced weighting mechanism:
\begin{equation}
    f^{av} = \sigma(f'^v U^v) + \sigma(f'^a U^a),
\end{equation}
where $U^v, U^a \in \mathbb{R}^{d \times d}$ are learnable projection matrices that transform the enhanced visual and audio features into a common representation space, and $f^{av} \in \mathbb{R}^{d}$ preserving the original feature dimension $d$.

\subsubsection{Hierarchical Representation Learning with Prototype-Enhanced Updates}

After obtaining the audio-visual fused representation $f_{av}$, we implement a hierarchical representation learning structure designed specifically for FFIA CIL. We detach the backbone and attach a 2-layer linear feed-forward network, where the first layer performs feature expansion before feeding to the subsequent layer for classification. The feature expansion process is formulated as:
\begin{equation}
F^{av} = Relu(f^{av}W_{up}^{av})
\end{equation}
where $f^{av}$ is the initial fused representation from our dual-encoder, $W_{up}^{av}$ is a randomly initialised and fixed weight matrix that projects features to a higher-dimensional space. This feature expansion significantly increases the feature dimension (typically by an order of magnitude), enhancing the model's representational capacity for analytical learning. Similarly, we obtain expanded representations $F^{v}$ and $F^{a}$ for visual and audio modalities, respectively.

Our hierarchical structure consists of two complementary components that work in tandem: a hierarchical representation framework that separates general intensity knowledge from species-specific characteristics, and a prototype management system that preserves discriminative information in feature space. Together, these components enable effective incremental learning across fish species without requiring storage of raw historical data.

\paragraph{Hierarchical Representation Framework}
 The general intensity layer is essential for cross-species knowledge transfer, allowing the model to recognize fundamental feeding intensity patterns regardless of the specific fish species. Meanwhile, the species-specific layers account for the unique behavioural manifestations of different fish species at the same intensity levels.

For the general intensity layer, we initially compute weights using a closed-form solution as follows:
\begin{equation}
W^{av} = ((F^{av})^{\top} F^{av} + \eta I)^{-1} (F^{av})^{\top} Y
\end{equation}
where $F^{av}$ is the matrix of up-sampled audio-visual features, $Y$ represents the intensity labels encoded as one-hot vectors, $I$ is the identity matrix, and $\eta$ is a regularisation parameter that prevents overfitting and ensures numerical stability, $\top$ indicates the transpose operation. This closed-form solution provides an optimal set of weights that minimises the regularised mean squared error without requiring iterative optimisation.

For each fish species introduced at step $k$, we compute modality-specific classifiers to capture their unique audio-visual signatures:
\begin{equation}
\begin{split}
W_{k}^a &= ((F_{k}^a)^{\top} F_{k}^a + \eta I)^{-1} (F_{k}^a)^{\top} Y_{k} \\
W_{k}^v &= ((F_{k}^v)^{\top} F_{k}^v + \eta I)^{-1} (F_{k}^v)^{\top} Y_{k}
\end{split}
\end{equation}
where $F_{k}^a \in \mathbb{R}^{n_k \times d_a}$ and $F_{k}^v \in \mathbb{R}^{n_k \times d_v}$ represent the audio and visual features from species $k$ with $n_k$ samples, $Y_{k} \in \mathbb{R}^{n_k \times c}$ contains the corresponding intensity labels, and $\eta$ is a regularization parameter. These classifiers are derived using the closed-form solution to the ridge regression problem (regularised least squares), allowing the model to recognise the distinctive acoustic and visual patterns associated with each fish species during feeding.

\paragraph{Prototype Management System}
Our prototype management system serves as a compact memory mechanism that preserves essential feeding intensity patterns without storing raw data. We maintain hierarchical prototype sets that capture both general intensity patterns and species-specific characteristics:
\begin{equation}
\begin{split}
P^{av} &= \{p_{i,j}^{av} | i \in I, j \in \{1,2,...,m\}\} \\
P^{sp} &= \{P^{sp}_k | k = 1,2,...,K\}
\end{split}
\end{equation}
where $P^{av}$ is a collection of general intensity prototypes that capture the common feeding intensity patterns shared across all fish species. $i$ indexes the intensity level: (``None", ``Weak",``Medium", and ``Strong"). $j$ indexes the prototype within each intensity level (typically 1 to 5). $m$ is the number of prototypes per intensity level. These general prototypes help the model maintain knowledge about how feeding intensities appear regardless of the specific fish species. $P^{sp}_k$ is a collection of species-specific prototypes that capture the unique behavioural characteristics of each fish species at different feeding intensities. For each fish species $k$, the set $P^{sp}_k$ contains prototypes that represent how that particular species exhibits different audio and visual feeding intensities.

We generate these prototypes through k-means clustering in the feature space, which identifies the most representative examples for each intensity level:
\begin{equation}
p_{i,j}^{av} = \frac{1}{|S_{i,j}|}\sum_{F^{av} \in S_{i,j}} F^{av}
\end{equation}
where $S_{i,j}$ represents features assigned to cluster $j$ for intensity level $i$. $|S_{i,j}|$ is the number of features in this cluster. $\sum_{F^{av} \in S_{i,j}} F^{av}$ represents the sum of all feature vectors in the set $S_{i,j}$. Through extensive experiments, we found empirically that storing just 5 prototypes per intensity level provides a reasonable balance between memory efficiency and representational power.

For species-specific prototypes, we perform clustering independently for each fish species to preserve their unique behavioural signatures in audio and visual modality:
\begin{equation}
\begin{split}
p_{i,k,j}^a = \frac{1}{|S_{i,k,j}|}\sum_{F^{a} \in S_{i,k,j}} F^{a} \\
p_{i,k,j}^v = \frac{1}{|S_{i,k,j}|}\sum_{F^{v} \in S_{i,k,j}} F^{v} 
\end{split}
\end{equation}
where $S_{i,k,j}$ represents features from fish species $k$ at intensity level $i$ assigned to cluster $j$.

\subsection{Prototype-Enhanced Learning}

After establishing the hierarchical representation structure and prototype management system, we present our approach for utilising these prototypes during incremental learning. This section details how prototypes are integrated into the weight update process when new fish species are introduced and how they contribute to the final prediction mechanism with dynamic modality balancing.

\subsubsection{Prototype-Enhanced Incremental Updates}
Rather than maintaining separate update mechanisms, we integrate prototypes directly into the weight update process to create a more unified and effective approach. At the incremental stage $k$ when a new fish species is introduced, we enhance the feature set by including prototypes from previous stages as additional training examples:
\begin{equation}
\begin{split}
\tilde{F}_k^{av} &= [F_k^{av}; \lambda_p \cdot P^{av}] \\
\tilde{Y}_k &= [Y_k; \text{OneHot}(I_{proto})]
\end{split}
\end{equation}
where $\tilde{F}_k^{av} \in \mathbb{R}^{(n_k+m \cdot |I|) \times d_{up}}$ represents the augmented feature matrix that includes both current fish species data and prototypes from previous stages. $F_k^{av} \in \mathbb{R}^{n_k \times d_{up}}$ is the feature matrix of the current fish species $k$ with $n_k$ samples. $P^{av} \in \mathbb{R}^{(m \cdot |I|) \times d_{up}}$ is a matrix of general intensity prototypes from previous species, where $m$ is the number of prototypes per intensity level (typically 5) and $|I|$ is the number of intensity levels (4 in our case). [;] represents a vertical concatenation operator (stacking matrices on top of each other). $I_{proto}$ represents Intensity level indices for each prototype (to which class each prototype belongs). $\tilde{Y}_k$ contains the corresponding labels, and $\lambda_p$ uses adjusted cosine similarity (max(0.2, similarity)) between new species features and existing prototypes, balancing historical knowledge preservation with adaptation to new species characteristics.

We then update the general intensity weights using this prototype-enhanced feature set:
\begin{equation}
W^{av}_k = ((\tilde{F}_k^{av})^{\top} \tilde{F}_k^{av} + \eta I)^{-1} (\tilde{F}_k^{av})^{\top} \tilde{Y}_k
\end{equation}

This prototype-enhanced update mechanism provides several advantages over traditional approaches. Directly incorporating semantic information from previous species into the weight update, facilitates more effective knowledge transfer across fish species. It maintains the computational efficiency of analytical solutions without requiring iterative optimisation or access to historical raw data. Most importantly, it creates a more unified framework for knowledge preservation by using the same prototype representation for both parameter updates and prediction enhancement.

During incremental learning, we also selectively update the general intensity prototypes to incorporate new knowledge while maintaining stability:
\begin{equation}
p_{i,j}^{new} = \alpha \cdot p_{i,j}^{old} + (1-\alpha) \cdot \frac{1}{|S_{i,j}^{new}|}\sum_{F^{av} \in S_{i,j}^{new}} F^{av}
\end{equation}
where $p_{i,j}^{new}$ is the updated prototype (Prototypes in our framework are centroid vectors derived from k-means clustering of feature representations, serving as compact memory units that preserve essential characteristics of each feeding intensity level across different fish species.), $p_{i,j}^{old}$ is the old prototype, $\frac{1}{|S_{i,j}^{new}|}\sum_{F^{av} \in S_{i,j}^{new}} F^{av}$ is the average vector of the new sample cluster, and $\alpha$ is a stability coefficient that prevents catastrophic forgetting while allowing limited adaptation to new patterns. Through extensive experimentation, we found that setting $\alpha = 0.7$ provides an optimal balance between stability and plasticity.

It is important to note that our framework employs two distinct weighting mechanisms that operate at different stages of the incremental learning process. While $\lambda_p$ controls how much influence existing prototypes have during the weight update process based on their similarity to new data, $\alpha$ governs how much the prototypes themselves change after learning from new data. This dual-weighting strategy enables our framework to adaptively utilize relevant historical knowledge during learning $(\lambda_p)$ while also ensuring controlled preservation of that knowledge after learning $(\alpha)$, creating a robust balance between stability and plasticity.

\subsubsection{Final Prediction with Dynamic Modality Balancing}
For prediction, we combine the outputs from both the general intensity and species-specific layers, with dynamic modality balancing to adapt to different feeding conditions:
\begin{equation}
\begin{split}
\hat{y}^{av}_k &= (W^{av}_k)^{\top} F^{av}_k \\
\hat{y}_k^{sp} &= \beta^a_{k,i} \cdot (W^a_k)^{\top} F^a_k + \beta^v_{k,i} \cdot (W^v_k)^{\top} F^v_k \\
\hat{y} &= \text{softmax}(\gamma_k \cdot \hat{y}_k^{av} + (1-\gamma_k) \cdot \hat{y}_k^{sp})
\end{split}
\end{equation}
where $\beta_{k,i}^a$ and $\beta_{k,i}^v$ are modality importance weights that adapt to different feeding stages and environmental conditions, and $\gamma_k$ is an adaptive balancing parameter that weights the contribution of general versus species-specific knowledge based on prediction confidence.

The modality importance weights are computed dynamically:
\begin{equation}
\beta_{k,i}^a = \sigma(w_{k,i}^{\top}[F_k^a, F_k^v])
\end{equation}
where $\sigma$ is the sigmoid function, $w_{k,i}$ is a learnable parameter vector specific to fish species $k$ and intensity level $i$ that weights the relative importance of modality features, and $[F_k^a, F_k^v]$ represents the concatenated audio and visual features. The visual modality weight is complementary: $\beta_{k,i}^v = 1 - \beta_{k,i}^a$.

The balancing parameter $\gamma_k$ decreases as more species are learned:
\begin{equation}
\gamma_k = \gamma_{\text{max}} - (\gamma_{\text{max}} - \gamma_{\text{min}}) \cdot \frac{k}{K}
\end{equation}
where $k$ is the current incremental stage, $K$ is the total number of stages, and $\gamma_{\text{max}}$ and $\gamma_{\text{min}}$ are the maximum and minimum values (typically set to 0.8 and 0.3, respectively). This causes the model to rely more on the species-specific layer in later stages, leveraging the more accurate species-specific representations that have been learned.

\section{Dataset}
\label{sec:dataset}

To address the scarcity of publicly available datasets for multimodal FFIA, we introduce the AV-CIL-FFIA dataset, a comprehensive and large-scale dataset designed specifically for advancing research in AV-CIL for FFIA.

\begin{figure}
\centering
\includegraphics[width=8cm]{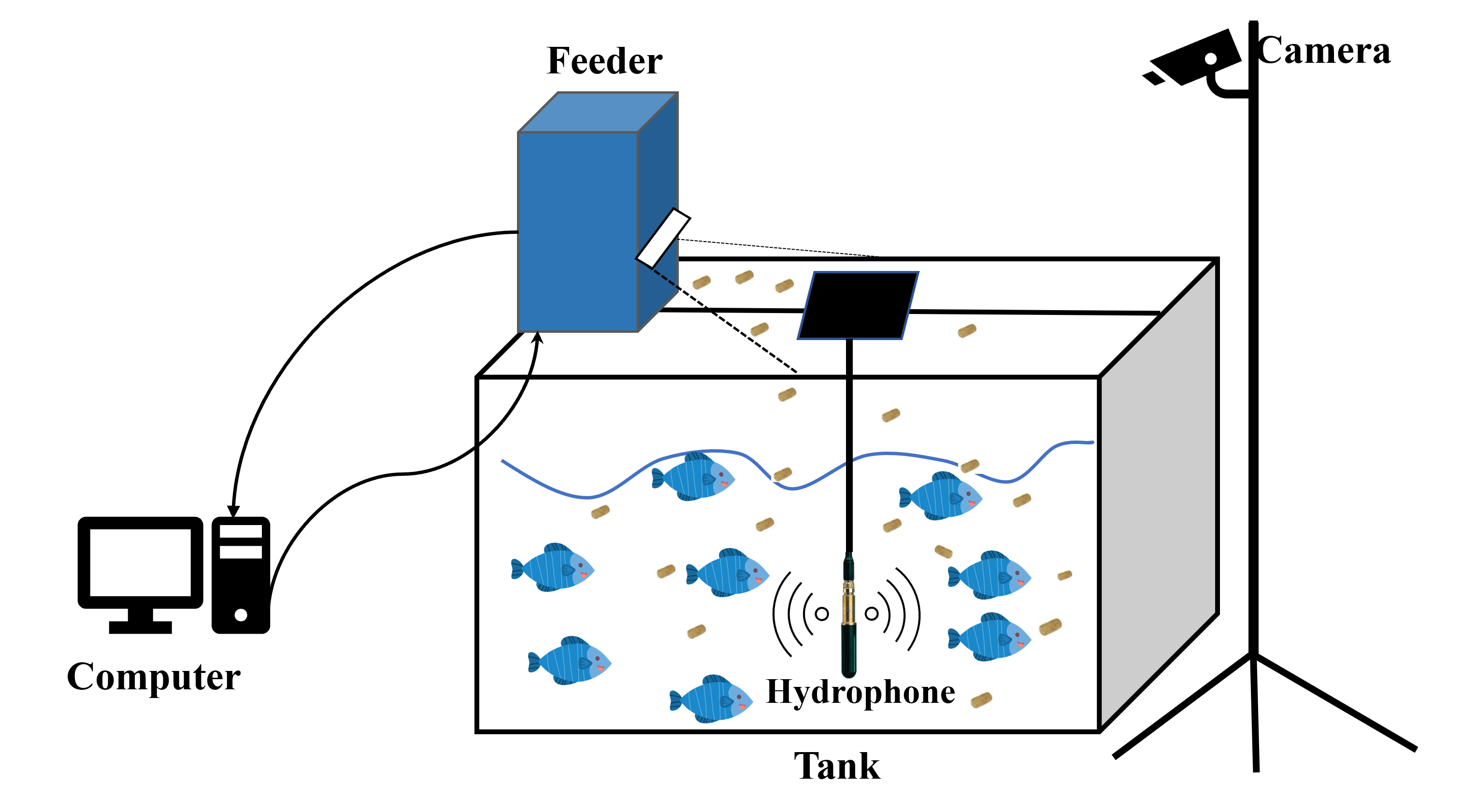}
\caption{Experimental systems for data collection. A hydrophone was underwater, and the camera was deployed on a tripod at a height of about two meters to capture the video data.}
\label{fig: experiment system}
\end{figure}

\subsection{Dataset Collection and Specifications}

The AV-CIL-FFIA dataset was collected in a real aquaculture environment at the Aquatic Products Promotion Station in Guangzhou, China. We collected a total of 81,932 audio-visual samples, focusing on six different fish species commonly farmed in aquaculture: Tilapia, Black Perch, Jade Perch, Lotus Carp, Red Tilapia, and Sunfish. The real-world setting introduced various challenges such as environmental noise, water surface reflection, and foam, providing a more realistic and complex scenario compared to controlled environments.

For data collection, we used six identical tanks, each measuring 4 meters in length, 2 meters in width, and 3 meters in depth. We employed a high-definition digital camera (Hikvision DS-2CD2T87E(D)WD-L) with a frame rate of 25 fps (1920 × 1080) and a high-frequency hydrophone (LST-DH01) with a sampling frequency of 256 kHz. As shown in Fig. \ref{fig: experiment system}, the camera was mounted on a tripod approximately 2 meters above the water surface to capture video data, while the hydrophone was submerged to record audio data. The acquisition of video and audio data was synchronised to ensure temporal alignment of the multimodal information.
Each sample in the dataset consists of a 2-second audio clip and its corresponding video clip. Experienced aquaculture technicians manually annotated the samples to provide ground-truth labels for model evaluation. The feeding intensity was categorized into four levels: ``Strong", ``Medium", ``Weak", and ``None", based on the observed feeding behaviour.

We randomly split the AV-CIL-FFIA dataset into training (70\%, 57,352 samples), validation (10\%, 8,193 samples), and testing (20\%, 16,387 samples) sets. This split ensures a robust evaluation of model performance across different fish species and feeding intensities.
The AV-CIL-FFIA dataset aims to bridge the gap in the current literature by providing a diverse and well-annotated collection of synchronised fish-feeding audio and video clips from multiple species in a real-world setting. This comprehensive dataset enables researchers to develop, compare, and evaluate novel multimodal FFIA methods, particularly in the context of class-incremental learning.

\subsection{Dataset Visualisation and Analysis}

\begin{figure*}[t]
\centering
\includegraphics[scale=1.2]{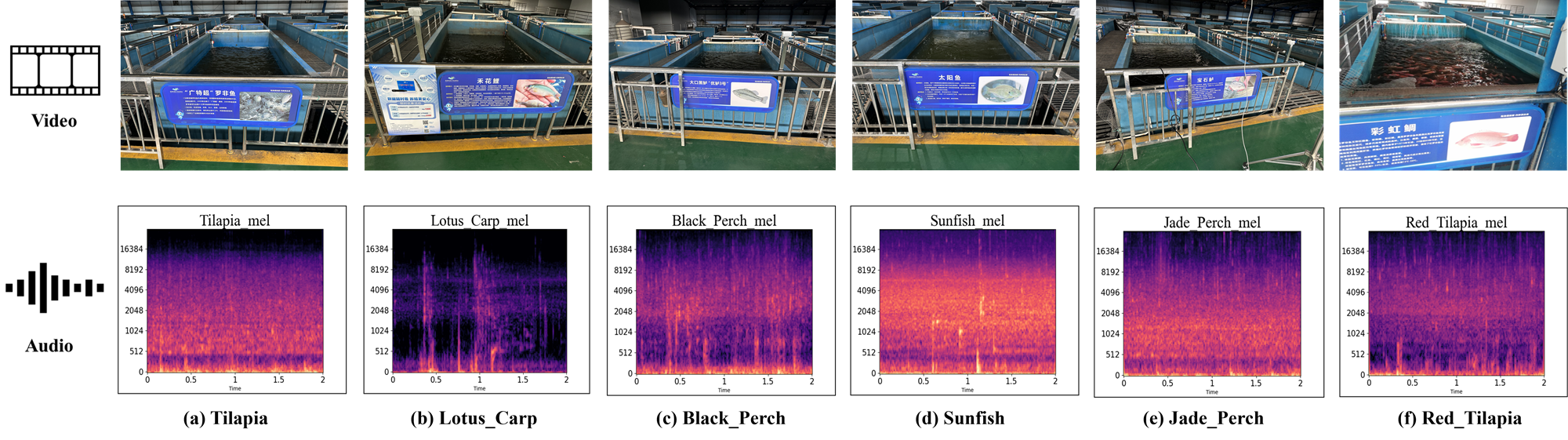}
\caption{Comparison of video frames (top) and audio mel-spectrograms (bottom) across the six fish species in the AV-CIL-FFIA dataset: (a) Tilapia, (b) Lotus Carp, (c) Black Perch, (d) Sunfish, (e) Jade Perch, and (f) Red Tilapia. The video frames demonstrate the challenging visual conditions in real aquaculture environments, while the mel-spectrograms reveal distinct acoustic signatures characteristic of each species during feeding.}
\label{fig: species_comparison}
\end{figure*}

To better understand the AV-CIL-FFIA dataset characteristics, Fig. \ref{fig: species_comparison} presents video frames and corresponding audio mel-spectrograms for all six fish species. The visual data reveals significant challenges for FFIA in real aquaculture environments. As evident from the images, water turbidity varies considerably between tanks, with most exhibiting cloudy conditions that make individual fish difficult to discern. Only certain species with distinctive colouration, such as Red Tilapia (Fig. \ref{fig: species_comparison}f), remain somewhat visible despite these challenging conditions.
The audio mel-spectrograms, however, reveal rich information that complements the limited visual data. Each fish species demonstrates distinctive acoustic signatures during feeding. Tilapia (Fig. \ref{fig: species_comparison}a) exhibits consistent energy distribution across multiple frequency bands, while Lotus Carp (Fig. \ref{fig: species_comparison}b) shows more concentrated energy in lower frequencies with distinct temporal patterns. Black Perch (Fig. \ref{fig: species_comparison}c) produces clear mid-frequency components, and Sunfish (Fig. \ref{fig: species_comparison}d) displays a broader energy distribution with characteristic patterns around 1024-2048 Hz. Jade Perch (Fig. \ref{fig: species_comparison}e) generates strong lower frequency components with periodic intensity variations, while Red Tilapia (Fig. \ref{fig: species_comparison}f) shows intense energy concentrations in the lower frequency range. While the acoustic and visual characteristics vary significantly between species, the underlying feeding intensity patterns share common features that can be leveraged for cross-species knowledge transfer.

\section{Experimental Setup}
\label{sec:experiment}

\subsection{Audio Data Processing}

We use the MobileNetV2 \cite{kong2020panns} model pretrained on AudioSet \cite{gemmeke2017audio} as our audio feature extractor. Following standard audio processing procedures, the original audio sampling rate of \num{256} kHz is downsampled to \num{64} kHz to reduce computational complexity. For feature extraction, we compute mel spectrograms with \num{128} mel bins using a Hanning window of \num{2048} samples and a hop size of \num{1024} samples. This results in a mel spectrogram with dimensions of \num{128}$\times$\num{128} for each \num{2}-second audio clip. During training, we apply SpecAugment \cite{park2019specaugment} for data augmentation, which masks blocks of consecutive frequency channels and time steps in the spectrogram.

\subsection{Video Data Processing}
For video processing, we use the S3D \cite{xie2018rethinking} model pretrained on Kinetics-400 \cite{kay2017kinetics} to extract visual features. From each \num{2}-second video clip with an original resolution of \num{2560}$\times$\num{1440}, we randomly sample \num{16} frames and resize them to \num{224}$\times$\num{224} pixels. This sampling strategy ensures that we capture sufficient temporal information while maintaining computational efficiency.

\subsection{Training Setup}
All experiments are conducted using PyTorch. For the initial training phase with the first fish species (Red\_Tilapia), we perform full fine-tuning of both encoders to adapt them to the aquaculture domain. For subsequent incremental learning stages, we follow a specific order of fish species: Tilapia, Jade\_Perch, Black\_Perch, Lotus\_Carp, and finally Sunfish. We freeze the pre-trained audio and visual encoders after the initial stage and only update the hierarchical representation layers using our prototype-enhanced approach.

The feature extraction backbone is initially trained using the Adam optimiser \cite{kingma2014adam} with a batch size of 64 samples. The learning rate is 1e-3 and the weight decay of 1e-4 for a maximum of 100 epochs.  However, for the incremental learning components, we employ analytical learning with closed-form solutions as shown in our formulation. For the regularisation parameter $\eta$ in our analytical learning formulation, we set it to 1.0 in all experiments. The prototype library maintains 5 representative vectors per intensity level for each species of fish, which provides an appropriate balance between memory efficiency and representation capacity. All experiments were performed using one NVIDIA RTX 3090 GPU with results averaged over 3 independent runs to ensure statistical reliability.

\subsection{Baseline Methods}

We evaluated our HAIL-FFIA framework against various state-of-the-art incremental learning methods. This comprehensive comparison includes both exemplar-based and exemplar-free approaches to demonstrate the effectiveness of our proposed method.
For exemplar-free approaches, we compare with: (1) Fine-tuning\footnotemark[1], which initializes the model with parameters trained from the previous task and retrains it on the current task without any mechanisms to prevent catastrophic forgetting; (2) LwF \footnote{ \url{https://github.com/weiguoPian/AV-CIL_ICCV2023}} \cite{li2017learning}, which employs knowledge distillation to preserve knowledge of previously learned tasks; (3) ACIL\footnote{ \url{https://github.com/ZHUANGHP/Analytic-continual-learning}}\cite{zhuang2022acil}, which uses analytical learning with closed-form solutions to avoid storing exemplars; and (4) MMAL \cite{yue2024mmal}, a recent multi-modal analytical learning approach that reformulates incremental learning from a recursive least squares perspective, offering an exemplar-free solution while addressing modality-specific knowledge compensation.

For exemplar-based methods, we compare with: (1) iCaRL\footnotemark[1] \cite{rebuffi2017icarl}, evaluated with both nearest-mean-of-exemplars classification (iCaRL-NME) and the fully connected classifier (iCaRL-FC); (2) SS-IL\footnotemark[1] \cite{ahn2021ssil}, which selects structurally significant samples as exemplars; (3) AFC\footnotemark[1] \cite{kang2022class}, tested with both nearest-mean-of-exemplars (AFC-NME) and the learned classifier (AFC-LSC); and (4) AV-CIL\footnotemark[1], which specifically addresses audio-visual incremental learning using class token distillation and modality-specific knowledge preservation.

To establish an upper performance bound, we report results from an Oracle model that has access to all training data from all fish species simultaneously, representing the theoretical maximum performance achievable. For a fair comparison, all baseline methods use the same backbone architecture as our approach. For exemplar-based methods, we adopt the exemplar selection strategies described in their original papers and ensure that all methods use the same memory budget for storing exemplars, making the comparison equitable.

\subsection{Evaluation Metrics}
Following the research in continual learning \cite{wang2022learning, pian2023audio}, two metrics, i.e., average accuracy and average forgetting, are used to evaluate the overall accuracy in continual learning stages and the average decrease of accuracy in previous tasks, respectively, which are defined as follows.

\subsubsection{Average accuracy (Avg Acc)} 
Here, $a_{k, j} \in[0,1]$ denotes the accuracy evaluated on the test set of task $j$ after learning task $k$ $(j \leq k)$. Then the average accuracy on the task $k$ can be calculated as:
\begin{equation}
A_k=\frac{1}{k} \sum_{j=1}^k a_{k, j}
\end{equation}

\subsubsection{Average forgetting (Forget)} 
The forgetting for a certain task is defined as the difference between the maximum knowledge obtained with respect to the task during the learning process in the past and the current knowledge the model has about it. $f_j^k \in[-1,1]$ denotes the forgetting on the previous task $j$ after learning task $k$, which can be formulated as:

\begin{equation}
f_j^k=\max_{l \in {j, \cdots, k-1}} a_{l,j}-a_{k,j}, \quad \forall j<k
\end{equation}
where $f_j^k$ represents the forgetting on previous task $j$ after learning task $k$, $a_{l, j}$ denotes the accuracy evaluated on task $j$ after learning task $l$, and $\max_{l \in {j, \cdots, k-1}} a_{l,j}$ identifies the maximum accuracy previously achieved on task $j$ during the incremental learning process.
Thus, the average forgetting at the $k$-th task can be defined as:

\begin{equation}
F_k=\frac{1}{k-1} \sum_{j=1}^{k-1} f_j^k
\end{equation}

Note that the lower the average forgetting $F_k$, the less forgetting of the model on the previous tasks.
\section{Results and Analysis}
\label{sec:results}

\subsection{The Results of Audio-Visual based CIL FFIA}

\begin{table*}
\centering
\small
\setlength{\tabcolsep}{4pt}
\newcolumntype{C}{>{\centering\arraybackslash}c}
\caption{The performance comparison of different audio-visual CIL methods and our proposed HAIL-FFIA}
\label{tab:AV result}
\begin{tabular}{@{}lCcccccc@{}}
\toprule
\multirow{2}{*}{Method} & \multirow{2}{*}{Exemplar-free?} & \multicolumn{2}{c}{Audio} & \multicolumn{2}{c}{Visual} & \multicolumn{2}{c}{Audio-Visual} \\ 
\cmidrule(lr){3-4} \cmidrule(lr){5-6} \cmidrule(lr){7-8}
                        &  & Avg Acc $\uparrow(\%)$ & Forget $\downarrow(\%)$ & Avg Acc $\uparrow(\%)$ & Forget $\downarrow(\%)$ & Avg Acc $\uparrow(\%)$ & Forget $\downarrow(\%)$ \\ 
\midrule
Fine-tuning              & \XSolidBrush  & 39.26 & 53.42 & 42.38 & 48.76 & 45.64 & 42.31 \\ 
LwF \cite{li2017learning}                  & \Checkmark & 45.31 & 45.23 & 48.75 & 42.39 & 52.83 & 36.25 \\ 
ACIL \cite{zhuang2022acil}                  & \Checkmark & 50.76 & 39.84 & 52.97 & 38.52 & 56.33 & 32.41 \\ 
iCaRL-NME \cite{rebuffi2017icarl}               & \XSolidBrush & 52.52 & 30.25 & 55.29 & 28.63 & 58.28 & 25.78 \\ 
iCaRL-FC \cite{rebuffi2017icarl}              & \XSolidBrush & 51.38 & 33.47 & 57.24 & 30.82 & 59.81 & 27.35 \\ 
SS-IL \cite{ahn2021ssil}                 & \XSolidBrush & 57.25 & 31.18 & 61.84 & 28.28 & 64.36 & 23.47 \\ 
AFC-NME \cite{kang2022class}                & \XSolidBrush & 56.42 & 29.84 & 60.27 & 27.53 & 63.19 & 24.86 \\ 
AFC-LSC \cite{kang2022class}                & \XSolidBrush & 62.32 & 23.79 & 65.16 & 21.85 & 68.93 & 17.72 \\ 
AV-CIL \cite{pian2023audio}                 & \XSolidBrush & 67.41 & 18.36 & 69.82 & 16.25 & 73.26 & 12.48 \\ 
MMAL \cite{yue2024mmal}                   & \Checkmark   & 59.24 & 26.32 & 61.53 & 24.76 & 64.28 & 21.54 \\ 
\textbf{HAIL-FFIA} (Ours)                    & \Checkmark   & \textbf{70.27} & \textbf{14.43} & \textbf{72.58} & \textbf{12.25} & \textbf{75.92} & \textbf{9.36} \\ 
\midrule
Oracle (Upper Bound)                  & \XSolidBrush   & 79.26 & - & 82.44 & - & 86.78 & - \\ 
\bottomrule
\end{tabular}
\end{table*}

Table \ref{tab:AV result} presents the comparative performance of our HAIL-FFIA framework against various state-of-the-art methods on the AV-CIL-FFIA dataset. The results are reported in terms of average accuracy (Avg Acc) and forgetting measure (Forget) across all six fish species for three different input modalities: audio-only, visual-only, and audio-visual.

\begin{figure}
\centering
\includegraphics[width=8cm]{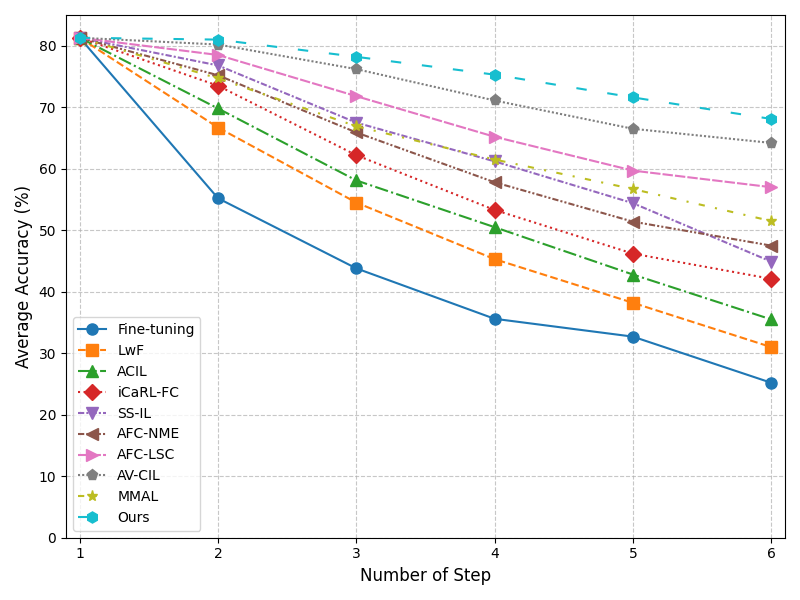}
\caption{Testing accuracy at each incremental step on AV-CIL-FFIA. The results show that as the incremental step increases, our methods generally outperform other state-of-the-art incremental learning methods.}
\label{fig: experiment figure}
\end{figure}

Our HAIL-FFIA framework consistently outperforms all baseline methods across all modalities. In the audio-visual setting, which leverages both modalities, our method achieves the highest average accuracy of 75.92\% and the lowest forgetting measure of 9.36\%, compared to the best-performing baseline, AV-CIL, with 73.26\% accuracy and 12.48\% forgetting. Fig. \ref{fig: experiment figure} illustrates how classification accuracy evolves as new fish species are incrementally added. All methods show a downward trend but with significantly different degradation rates. Fine-tuning suffers the most severe forgetting, dropping from 81.3\% to approximately 25\% by the final step, as it lacks mechanisms to preserve previous knowledge. Knowledge distillation methods like LwF perform better (31.6\% at the final step) by transferring knowledge from previous models, but still struggle with the stability-plasticity trade-off.

Exemplar-based methods (iCaRL, SS-IL, AFC, AV-CIL) show improved performance compared to exemplar-free approaches, with AV-CIL reaching 73.26\% average accuracy. These methods benefit from storing and replaying actual samples from previous species, but this comes at the cost of substantial storage requirements. AV-CIL particularly excels by explicitly modeling the relationships between audio and visual modalities but still requires storing raw data from all previously encountered species. Among exemplar-free approaches, MMAL (64.28\%) demonstrates good performance through analytical learning, but it lacks the hierarchical structure and dynamic modality balancing needed to fully adapt to the unique characteristics of different fish species during feeding. 

The performance advantages of our HAIL-FFIA method can be attributed to several key factors. The hierarchical representation effectively separates general intensity patterns from species-specific characteristics, allowing better knowledge transfer across different fish species without interference. When a new species is introduced, the general intensity layer captures common feeding behaviours while the species-specific layer adapts to unique characteristics, maintaining a balance between stability and plasticity. Additionally, our prototype-enhanced parameter updates preserve essential information in a highly compressed form without storing raw data, addressing the core challenge of exemplar-free learning. The dynamic modality balancing mechanism further enhances performance by adaptively adjusting the importance of audio versus visual information based on feeding stages and environmental conditions, a critical advantage in aquaculture settings where water turbidity and ambient noise can vary significantly.
The performance decrease observed at the final step (Sunfish) in all methods aligns with our dataset characteristics, as Sunfish samples have lower quality due to health problems in that population. Yet, even under these challenging conditions, HAIL-FFIA demonstrates superior robustness and adaptation capability.

\subsection{Comparison with Single Modality Based Results}

To evaluate the effectiveness of audio-visual fusion in class-incremental learning for FFIA, we compare the performance of all methods using single-modality (audio-only or visual-only) versus multi-modal inputs. As shown in Table \ref{tab:AV result}, training with joint audio-visual modalities consistently achieves higher average accuracy and lower forgetting in all methods compared to single-modality based approaches.
For our HAIL-FFIA method, the audio-visual configuration achieves 75.92\% average accuracy, outperforming audio-only (70.27\%) and visual-only (72.58\%) approaches by 5.65\% and 3.34\% respectively. The forgetting measure also shows substantial improvement with audio-visual integration (9.36\%) compared to single-modality configurations (14.43\% for audio-only and 12.25\% for visual-only). Similar patterns are observed across all baseline methods, confirming that multi-modal learning provides inherent advantages for class-incremental learning in FFIA tasks.

The performance advantage of multi-modal learning is particularly evident in the later incremental steps, suggesting that integrating audio and visual information provides greater resilience against catastrophic forgetting as more fish species are added. These results demonstrate that audio and visual modalities offer complementary information about feeding intensity, with audio signals better capturing certain feeding behaviours, while visual signals excel in others. Our HAIL-FFIA framework effectively leverages these complementary strengths through its dynamic modality balancing mechanism, enhancing both accuracy and knowledge retention in class-incremental learning scenarios.

\begin{table}[htbp]
  \centering
  \caption{Ablation study on our HAIL-FFIA framework. Our full model effectively mitigates catastrophic forgetting and achieves the best incremental learning performance on the AV-CIL-FFIA dataset.}
    \scalebox{0.9}{\begin{tabular}{ccccc}
    \toprule
    \multirow{10}[6]{*}{HAIL-FFIA} & \multicolumn{2}{c}{Representation Structure} & \multirow{2}[4]{*}{DMB} & \multirow{2}[4]{*}{Mean Acc.} \\
\cmidrule{2-3}          & HR & PM &       &  \\
\cmidrule{2-5}          & \XSolidBrush     & \XSolidBrush     & \XSolidBrush     & 50.25 \\
          & \CheckmarkBold     & \XSolidBrush     & \XSolidBrush     & 63.83 \\
          & \XSolidBrush     & \CheckmarkBold     & \XSolidBrush     & 61.42 \\
          & \XSolidBrush     & \XSolidBrush     & \CheckmarkBold     & 56.75 \\
          & \CheckmarkBold     & \CheckmarkBold     & \XSolidBrush     & 68.46 \\
          & \CheckmarkBold     & \XSolidBrush     & \CheckmarkBold     & 65.67 \\
          & \XSolidBrush     & \CheckmarkBold     & \CheckmarkBold     & 63.18 \\
          & \CheckmarkBold     & \CheckmarkBold     & \CheckmarkBold     & 75.92 \\
    \bottomrule
    \end{tabular}%
    }
  \label{tab:ablation}%
\end{table}%

\begin{figure}
\centering
\includegraphics[width=8cm]{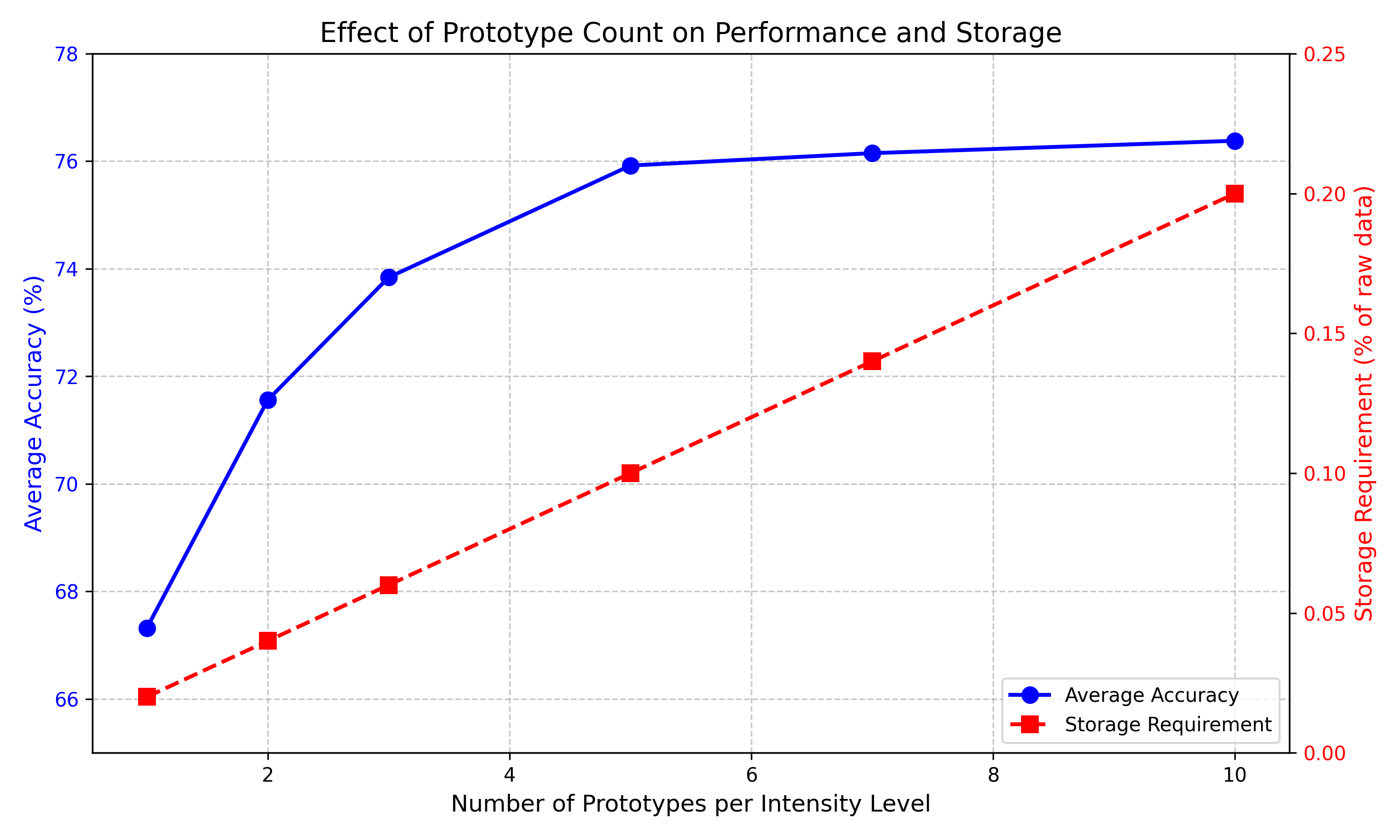}
\caption{Effect of prototype count on model performance and storage requirements. The blue line (left y-axis) shows average accuracy across all fish species, while the red line (right y-axis) indicates storage requirements as a percentage of raw data. Five prototypes per intensity level provide an optimal balance between accuracy and memory efficiency.}
\label{fig: prototype figure}
\end{figure}

\subsection{Ablation Studies}

To validate the effectiveness of each component in our HAIL-FFIA framework, we conducted comprehensive ablation studies, examining their individual and combined impacts on performance.

\subsubsection{Impact of Key Components}
Table \ref{tab:ablation} presents the results of our ablation study, where we systematically evaluate all possible combinations of our three key components: Hierarchical Representation (HR), Prototype Management (PM), and Dynamic Modality Balancing (DMB).

The baseline model without any of these components achieves 50.25\% mean accuracy. Adding the Hierarchical Representation structure alone improves performance to 63.83\%, making it the most impactful individual component. This confirms that separating general intensity patterns from species-specific characteristics significantly enhances knowledge transfer between fish species. The Prototype Management component contributes the second-largest individual improvement (61.42\%), demonstrating the value of our prototype-based knowledge preservation approach. The Dynamic Modality Balancing mechanism provides a more modest but still meaningful improvement (56.75\%) when added independently.

When combining components, we observe clear synergistic effects. The combination of Hierarchical Representation and Prototype Management yields 68.46\% accuracy, substantially higher than either component alone. The complete model with all three components achieves the best performance at 75.92\%, showing that each element makes a valuable contribution to the overall framework.

\subsubsection{Effect of Prototype Count}
We also investigated how the number of prototypes per intensity level affects the performance and memory efficiency of our model. Fig. \ref{fig: prototype figure} shows the trade-off between average accuracy and storage requirements as the prototype count varies from 1 to 10 per intensity level per fish species.

With just a single prototype per intensity level, the model achieves only 67.32\% average accuracy, demonstrating that a single prototype is insufficient to capture the variability within each intensity class. Increasing to 2 prototypes improves accuracy to 71.56\%, while 3 prototypes further enhance performance to 73.84\%. Our default configuration of 5 prototypes achieves 75.92\% accuracy, representing an optimal balance between performance and memory efficiency.
Further increasing to 7 or 10 prototypes yields diminishing returns (76.15\% and 76.38\% respectively), while storage requirements continue to grow linearly. This indicates that 5 prototypes per intensity level effectively capture the essential characteristics of feeding behaviours for each fish species while maintaining minimal storage requirements (approximately 0.10\% of the size of the original dataset).
For comparison, exemplar-based methods, such as iCaRL and AV-CIL, typically require 5-10\% of the original dataset, highlighting the substantial memory efficiency of our approach. This makes HAIL-FFIA particularly suitable for resource-constrained aquaculture monitoring systems that must be adapted to multiple fish species over time.
These ablation studies confirm that each component of our HAIL-FFIA framework makes a substantial contribution to overall performance and that our prototype-based approach offers an excellent balance between accuracy and memory efficiency for class-incremental learning in FFIA.

\section{Conclusion and Future Work}
\label{sec:conclusion}

In this paper, we have introduced HAIL-FFIA, a novel hierarchical audio-visual class-incremental learning framework for Fish Feeding Intensity Assessment that effectively adapts to new fish species while preserving knowledge of previously learned ones. Our approach employs a dual-layer hierarchical representation with prototype-enhanced parameter updates and dynamic modality balancing, enabling efficient incremental learning without storing raw data.
We also presented AV-CIL-FFIA, a comprehensive dataset comprising 81,932 labelled audio-visual clips across six fish species in real aquaculture environments. Our experimental results demonstrated that HAIL-FFIA significantly outperforms existing methods, achieving 75.92\% average accuracy with only 9.36\% forgetting—substantially better than both exemplar-free and exemplar-based approaches.
Through ablation studies, we verified the contribution of each component in our framework and established that five prototypes per intensity level provide an appropriate balance between performance and memory efficiency. Our multi-modal approach consistently outperformed single-modality configurations, confirming the benefits of integrating audio and visual information in FFIA tasks.
The prototype-based memory system of HAIL-FFIA requires just 0.1\% of the storage needed for raw data, making it exceptionally efficient for resource-constrained aquaculture monitoring systems. Future work will explore extending our approach to other aquaculture tasks, investigating more sophisticated prototype selection strategies, and developing deployment-optimized versions for commercial settings.

\section{ACKNOWLEDGMENT}
This work was supported by the National Natural Science Foundation of China “Dynamic regulation mechanism of nitrogen in industrial aquaponics under the condition of asynchronous life cycle [Grant no. 32373186], Digital Fishery Cross-Innovative Talent Training Program of the China Scholarship Council (DF-Project) and a Research Scholarship from the China Scholarship Council. For the purpose of open access, the authors have applied a Creative Commons Attribution (CC BY) licence to any author-accepted manuscript version arising.
\begin{bibliography}{mybib}
\bibliographystyle{ieeetr}
\end{bibliography}
\vspace{-40pt}
\end{document}